\documentclass[sigconf]{acmart}

\usepackage{graphicx}
\usepackage{amsbsy,amsmath,amsfonts}  
\usepackage{booktabs, booktabs}
\usepackage{multirow}
\usepackage{subfigure}
\usepackage{xcolor}
\usepackage{color,comment,ulem,soul}
\usepackage{algpseudocode}
\allowdisplaybreaks
\usepackage{algorithm}
\usepackage{makecell}

\newtheorem{thm}{Theorem}

\newtheorem{defn}[thm]{Definition}

\DeclareMathOperator{\trace}{trace}
\newcommand{\mc}{\mathcal}
\newcommand{\m}[1]{\mathbf{#1}}

\renewcommand{\mc}[1]{\ensuremath{\mathcal{#1}}} 	
\newcommand{\g}[1]{\mbox{\boldmath $#1$}}
\newcommand{\mb}[1]{{\mathbb{#1}}}

\AtBeginDocument{%
  }

\copyrightyear{2025}
\acmYear{2025}
\setcopyright{licensedusgovmixed}\acmConference[BCB '25]{Proceedings of the 16th ACM International Conference on Bioinformatics, Computational Biology, and Health Informatics}{October 11--15, 2025}{Philadelphia, PA, USA}
\acmBooktitle{Proceedings of the 16th ACM International Conference on Bioinformatics, Computational Biology, and Health Informatics (BCB '25), October 11--15, 2025, Philadelphia, PA, USA}
\acmDOI{10.1145/3765612.3767215}
\acmISBN{979-8-4007-2200-4/2025/10}

\begin{document}

\title{Fair CCA for Fair Representation Learning: An ADNI Study}

\author{Bojian Hou}
\authornote{Equal contributions}
\email{bojianh@upenn.edu}
\affiliation{%
  \institution{University of Pennsylvania}
  \city{Philadelphia}
  \state{Pennsylvania}
  \country{USA}
}

\author{Zhanliang Wang}
\authornotemark[1]
\email{aaronwzl@sas.upenn.edu}
\affiliation{%
  \institution{University of Pennsylvania}
  \city{Philadelphia}
  \state{Pennsylvania}
  \country{USA}
}

\author{Zhuoping Zhou}
\email{zhuopinz@sas.upenn.edu}
\affiliation{%
  \institution{University of Pennsylvania}
  \city{Philadelphia}
  \state{Pennsylvania}
  \country{USA}
}

\author{Boning Tong}
\email{boningt@seas.upenn.edu}
\affiliation{%
  \institution{University of Pennsylvania}
  \city{Philadelphia}
  \state{Pennsylvania}
  \country{USA}
}

\author{Zexuan Wang}
\email{zxwang@sas.upenn.edu}
\affiliation{%
  \institution{University of Pennsylvania}
  \city{Philadelphia}
  \state{Pennsylvania}
  \country{USA}
}

\author{Jingxuan Bao}
\email{bao96@sas.upenn.edu}
\affiliation{%
  \institution{University of Pennsylvania}
  \city{Philadelphia}
  \state{Pennsylvania}
  \country{USA}
}

\author{Duy Duong-Tran}
\email{dduongtr@upenn.edu}
\affiliation{%
  \institution{University of Pennsylvania}
  \city{Philadelphia}
  \state{Pennsylvania}
  \country{USA}
}

\author{Qi Long}
\email{qlong@pennmedicine.upenn.edu}
\affiliation{%
  \institution{University of Pennsylvania}
  \city{Philadelphia}
  \state{Pennsylvania}
  \country{USA}
}

\author{Li Shen}
\authornote{Corresponding author}
\email{li.shen@pennmedicine.upenn.edu}
\affiliation{%
  \institution{University of Pennsylvania}
  \city{Philadelphia}
  \state{Pennsylvania}
  \country{USA}
}

\renewcommand{\shortauthors}{Hou, Wang, et al.}

\begin{abstract}
  Canonical correlation analysis (CCA) is a technique for finding correlations between different data modalities and learning low-dimensional representations. As fairness becomes crucial in machine learning, fair CCA has gained attention. However, previous approaches often overlook the impact on downstream classification tasks, limiting applicability. We propose a novel fair CCA method for fair representation learning, ensuring the projected features are independent of sensitive attributes, thus enhancing fairness without compromising accuracy. We validate our method on synthetic data and real-world data from the Alzheimer's Disease Neuroimaging Initiative (ADNI), demonstrating its ability to maintain high correlation analysis performance while improving fairness in classification tasks. Our work enables fair machine learning in neuroimaging studies where unbiased analysis is essential. Code is available in https://github.com/ZhanliangAaronWang/FR-CCA-ADNI.
  
\end{abstract}




\maketitle

\section{Introduction}

Canonical correlation analysis (CCA) is a well-established technique for exploring the relationship between two sets of multi-dimensional variables. By finding a pair of linear transformations that maximize the correlation between the projected variables, CCA enables the discovery of latent factors that capture the shared information across different modalities~\cite{hotelling1992relations}. 
CCA’s ability to preserve cross-modal correlations has made it widely used in
biology \cite{cca_bio}, neuroscience \cite{cca_brain} to medicine \cite{cca_eeg}, and engineering \cite{cca_fault}.


\begin{figure*}
    \centering
    \includegraphics[width=1\textwidth]{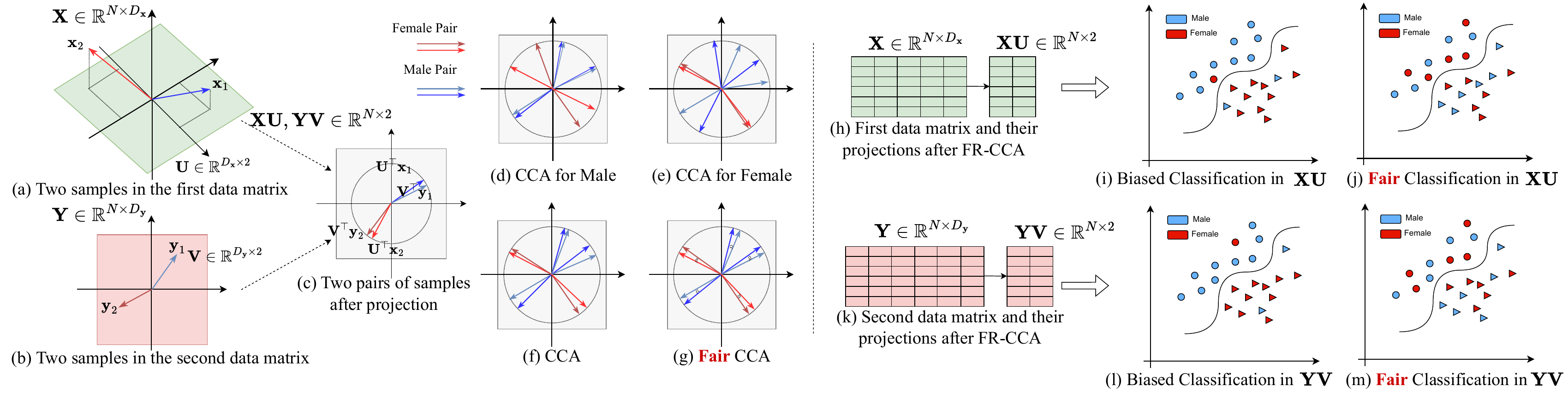}
    \caption{\label{fig:illustration}\small Illustration of FR-CCA with the sensitive attribute sex as an example (female and male). (a)--(c) demonstrate the general framework of CCA, while (d)--(g) provide a comparison of the projected results using various strategies. 
    It is important to note that the correlation between two corresponding samples is inversely associated with the angle formed by their projected vectors. 
    FR-CCA aims to equalize the average angles among different groups. After projection ((h) and (k)), our FR-CCA can lead to fair classification ((j) and (m)), where the classification results will not be affected by the sex information.}
\end{figure*}

Despite widespread adoption, conventional CCA methods do not account for potential biases concerning sensitive attributes such as sex, race, or age. This lack of fairness consideration can lead to learned representations that capture and amplify undesirable societal biases, resulting in discriminatory outcomes in downstream applications~\cite{barocas2019fairness,tarzanagh2023fairness}. As machine learning models are increasingly deployed in high-stakes decision-making scenarios, it is imperative to develop fair representation learning techniques that mitigate these biases while preserving the utility of the learned features~\cite{zemel2013learning}.


A recent study~\cite{zhou2024fair} developed fair CCA methods that minimize correlation disparity errors associated with protected attributes, ensuring equitable correlation levels across different groups. The authors proposed a single-objective fair CCA (SF-CCA) and a multi-objective fair CCA (MF-CCA). SF-CCA is more efficient with a fixed set of hyperparameters, while MF-CCA can automatically identify the optimal hyperparameters. However, these fair CCA methods primarily focus on correlation analysis without explicitly considering the subsequent classification or prediction tasks that often motivate using CCA. Such disconnection between the state-of-the-art CCA learnings and the downstream tasks could significantly hinder the broader applicability of fair CCA. Furthermore, although MF-CCA has the advantage of automatic hyperparameter tuning, its complex optimization techniques limit its efficiency, which, in turns, prevent its wider adoption in application domain.


In this paper, we introduce a novel fair CCA method that implicitly optimizes for both fairness and classification performance. Our approach enforces independence between the learned representations and sensitive attributes, thereby promoting fairness. 
By utilizing CCA, we maintain high correlation between multimodal data, which facilitates improved performance in subsequent classification tasks.

Specifically, as illustrated in Figure~\ref{fig:illustration}, panels (a)--(c) showcase the general framework of CCA. Two samples $\m{x}_1$ and $\m{x}_2$ from the first data matrix $\m{X}$, and $\m{y}_1$ and $\m{y}_2$ from the second data matrix $\m{Y}$ are projected into the same feature space. This projection aims to maximize correlations between corresponding samples, which are inversely related to the angles between the sample vectors $\m{x}_i$ and $\m{y}_i$ ($i=1 \text{ or } 2$).
Panels (d)--(g) compare the results of different learning strategies. Four pairs of samples are shown, with female pairs highlighted in red and male pairs in blue. Sex-based projection matrices heavily bias the final projection, favoring one sex over the other (panels (d) and (e)). While CCA reduces angles compared to random projection (panel (f)), significant angle differences between male and female pairs indicate persistent bias. Our proposed FR-CCA addresses this bias by maximizing correlation within pairs while ensuring equal correlations across different groups, such as males and females (panel (g)).
Panels (h) and (k) illustrate the original data matrices $\m{X}$ and $\m{Y}$ and their projections after FR-CCA, $\m{XU}$ and $\m{YV}$, respectively. Panels (i) and (j) demonstrate biased and fair classification on $\m{XU}$ before and after FR-CCA. The biased classification significantly relies on sex information, while fair classification shows more balanced sex representation in each class. Similar results are shown in panels (l) and (m) for $\m{YV}$ before and after FR-CCA.

We evaluate the proposed fair CCA method on various datasets, including synthetic and real datasets, with a particular focus on Alzheimer's disease (AD) diagnosis. In the clinical context of AD, ensuring fairness is crucial due to the demographic diversity of the patient population and the high stakes of diagnostic decisions. Traditional diagnostic models may inadvertently incorporate biases related to age, sex, or race, potentially leading to unequal access to diagnosis and treatment options. 

Applying our fair CCA method to AD diagnosis involves leveraging multimodal data, such as PET and MRI imaging information, to uncover latent factors critical for accurate diagnosis. By ensuring that these latent factors are unbiased regarding sensitive attributes, our method helps develop equitable diagnostic tools that are equitable across different demographic groups. This fairness aspect is particularly important in clinical settings, where biased diagnostic outcomes can significantly affect implications for patient care and treatment strategies \cite{doi:10.1126/science.aax2342}.

Our empirical results demonstrate that the proposed method achieves superior fairness-accuracy trade-offs compared to existing fair CCA techniques. In the clinical context of AD, this means more accurate and equitable diagnosis across diverse patient groups, enhancing the reliability and inclusiveness of diagnostic tools. By addressing both fairness and performance, our method offers a practical solution for fair representation learning in real-world scenarios, particularly in high-stakes fields like medicine, where equitable treatment is paramount.



\section{Related Work}

\textit{Canonical Correlation Analysis}
CCA, introduced by Hotelling in 1936~\cite{hotelling1936relations}, has been widely applied across various domains. In the context of representation learning, particularly for cross-modal applications, Deep CCA~\cite{andrew2013deep} extended traditional CCA to learn nonlinear transformations of two views of data, enabling the discovery of complex correlations in high-dimensional spaces. This advancement has been particularly relevant in fields such as neuroscience~\cite{cca_brain} and medicine~\cite{cca_eeg}, where multimodal data analysis \cite{nguyen2024volume} is crucial. Recent work has also explored probabilistic formulations of CCA~\cite{bach2005probabilistic}, further expanding its utility in domains like bioinformatics and computer vision.

\textit{Fairness in Machine Learning}
As machine learning models are increasingly deployed in high-stakes decision-making scenarios, ensuring fairness has become a critical concern. Seminal work by Dwork et al. (2012) \cite{dwork2012fairness} introduced the concept of individual fairness, while subsequent research has explored various group fairness metrics and their implications \cite{barocas2019fairness,tarzanagh2023fairness,hou2024pferm}. In the context of representation learning, Zemel et al. (2013)~\cite{zemel2013learning} proposed methods for learning fair representations independent of protected attributes. This line of research is particularly relevant to our work, as we aim to develop fair CCA methods that mitigate biases while preserving the utility of learned features.

\textit{Fair CCA}
Recent efforts have focused on incorporating fairness considerations directly into the CCA framework. Zhou et al. (2024)~\cite{zhou2024fair} developed fair CCA methods that minimize correlation disparity errors associated with protected attributes. Their work introduced single-objective fair CCA (SF-CCA) and multi-objective fair CCA (MF-CCA), to ensure equitable correlation levels across different groups. However, these methods primarily focus on the correlation analysis without explicitly considering for downstream tasks. Our work builds upon these foundations, addressing the limitations by jointly optimizing for fairness and classification performance.

\textit{Machine Learning for Alzheimer's Disease (AD)}
The application of machine learning to Alzheimer's Disease diagnosis has seen significant advancements in recent years. Multimodal approaches integrating neuroimaging, genetic, and clinical data have enhanced predictive performance~\cite{zhang2011multimodal,xu2022consistency,xu2024topology,xucaudal}. In the context of fair machine learning, recent work has highlighted the importance of developing unbiased diagnostic tools that perform equitably across diverse patient groups~\cite{doi:10.1126/science.aax2342}. Our research contributes to this area by applying fair CCA methods to AD diagnosis, leveraging multimodal data such as PET and MRI imaging information to uncover latent factors that are both critical for accurate diagnosis and unbiased with respect to sensitive attributes.


\section{Preliminaries}
\label{sec:pre}
CCA is a multivariate statistical technique that explores the relationship between two sets of variables \cite{hotelling1992relations}. Given two datasets $\m{X} 
\in \mb{R}^{N \times D_\m{x}}$ and  $\m{Y} \in  \mb{R}^{N \times D_\m{y}} $ on the same set of $N$ observations,
CCA seeks the $R$--dimensional subspaces where the projections of $\m{X}$ and $\m{Y}$ are maximally correlated. In other words, CCA finds $\m{U} \in  \mb{R}^{D_\m{x} \times R}$ and $\m{V} \in  \mb{R}^{D_\m{y} \times R }$ such that 
\begin{equation}\label{eqn:main:cca} \tag{CCA}
\begin{aligned}
\max & \quad \trace \left(\m{U}^\top\m{X}^\top\m{Y}\m{V}\right) \\
\textnormal{subject to} & \quad  \m{U}^\top\m{X}^\top \m{X}\m{U}=\m{V}^\top\m{Y}^\top\m{Y}\m{V} = \m{I}_R.
\end{aligned}
\end{equation}
Next, we review some notions of fairness in the classification task.
\begin{defn}\label{def:fair:notions}
Let $f$ be a score function that maps the random variable $X$ to a real number.
\begin{itemize}
\item \textnormal{ \textbf{Demographic Parity (DP)}}: $f$ satisfies demographic parity with respect to $A$ if $\ \mathbb{E}[f(X)] = \mathbb{E}[f(X)|A]$. 
\item  \textnormal{ \textbf{Equalized Odds (EO)}}: $f$ satisfies equalized odds with respect to $A$ if $\ \mathbb{E}[f(X)|Y] = \mathbb{E}[f(X)|Y,A]$. 
\item \textnormal{\textbf{Group Sufficiency (GS)}}: $f$ is sufficient with respect to attribute $A$ if $\ \mathbb{E}[Y|f(X)] = \mathbb{E}[Y|f(X), A]$.
\end{itemize}
\end{defn}

DP ensures that the expected score $f(X)$ remains constant, regardless of the attribute $A$. This principle guarantees that the distribution of scores remains unaffected by the sensitive attribute, thereby promoting fairness in the decision-making process. EO dictates that the expected score $f(X)$ remains consistent across all combinations of labels $Y$ and attributes $A$. It ensures that individuals sharing the same label but differing attributes are treated equally in terms of their predicted scores, irrespective of the sensitive attribute. GS ensures the score function $f$ captures all the information about the label $Y$ that is relevant for prediction, regardless of the attribute $A$.  
Definition~\ref{def:fair:notions} leads to a notion of the {\it demographic parity gap} (DPG), {\it equalized odds gap} (EOG), and {\it group sufficiency gap} (GSG) defined, respectively, as:
\begin{align}
&\textbf{DPG}_f(A) = \mb{E}[\mb{E}[f(X)] - \mb{E}[f(X)|A]], \\
&\textbf{EOG}_f(A) = \mb{E}[\mb{E}[f(X)|Y] - \mb{E}[f(X)|Y,A]], \\
&\textbf{GSG}_f(A) = \mb{E}[\left|\mb{E}[Y\mid f(X)]-\mb{E}[Y\mid f(X), A]\right|].
\end{align}

\section{Method}
The existing fairness notion in CCA~\cite{zhou2024fair} ensures that the local models learned from each group maintain a consistent distance from the global model in terms of correlation. However, this does not directly guarantee fairness for subsequent tasks, such as fairness-aware classification using notions such as demographic parity (DP), equalized odds (EO) or group sufficiency (GS)~\cite{donini2018empirical,shui2022learning,tarzanagh2023fairness}. This is because these fairness notions require that the classification predictions should not be influenced by the sensitive attribute, whereas previous fair CCA is not optimized for that purpose.
To satisfy this point, an appealing approach is {\it fair representation learning}~\cite{zemel2013learning}. Let $\m{x}\in\mc{X}$ denote a random vector representing features based on which predictions are made. 
The idea of fair representation learning is to learn a {fair} feature representation $f:\mathcal{X}\rightarrow \mathcal{X}'$ where $\m{x}$ is mapped to a new feature space $f(\m{x})\in\mathcal{X}'$ such that $f(\m{x})$ is (approximately) independent of the sensitive attribute.
Once a fair representation is found, any model trained on this representation will also be fair in terms of fairness notion in classification, such as DP, EO, or GS. Of course, the representation still needs to contain some information about $\m{x}$ in order to be useful. 

In our proposed fair CCA-based representation learning (FR-CCA), we remove sensitive information when projecting the dataset onto the $R$-dimensional linear subspace. We look for a best approximating projection such that the projected data does not contain sensitive information anymore: let $\m{z}\in\{0,1\}^N$ denote the vector containing the sensitive attribute of $N$ data points and let $z_i\in\{0,1\}$ denote its $i$th component corresponding to the $i$th data point, where the binary value encodes membership in one of two demographic groups and let $\m{X}_i\in\mathbb{R}^{1\times D_\m{x}}$ and $\m{Y}_i\in\mathbb{R}^{1\times D_\m{y}}$ denote the $i$th row of $\m{X}\in\mathbb{R}^{N\times D_\m{x}}$ and $\m{Y}\in\mathbb{R}^{N\times D_\m{y}}$ which represent the $i$th data point from two datasets separately.

Ideally, we aim to prevent any classifier from predicting $ z_i $ based solely on the projection of $\m{X}_i$ or $\m{Y}_i$ onto the $R$-dimensional subspace. Our goal is to solve the following:
\begin{equation}\label{eq:fair_CCA}
\begin{split}
\max_{\m{U},\m{V}}  \quad & \trace\left(\m{U}^\top\m{X}^\top\m{Y}\m{V}\right) \qquad
\text{subject to}  \quad \m{U}\in\mathcal{U},\m{V}\in\mathcal{V}, \\
\text{where}\  
& \mathcal{U}:=\left\{\m{U}\in\mathbb{R}^{D_\m{x}\times R}\mid\m{U}^\top\m{X}^\top\m{X}\m{U}=\m{I}_R;\quad\forall f:\mathbb{R}^R\to \mathbb{R},\right. \\  
&\left. \quad f(\m{U}^\top\m{X}_i^\top) ~\text{and}~ z_i \ \text{are statistically independent} \right\}, \\
&\mathcal{V}:=\left\{\m{V}\in\mathbb{R}^{D_\m{y}\times R}\mid\m{V}^\top\m{Y}^\top\m{Y}\m{V}=\m{I}_R;\quad \forall g:\mathbb{R}^R\to\mathbb{R},\right. \\ 
& \left.\quad g(\m{V}^\top\m{Y}_i^\top) ~\text{and}~ z_i \ \text{are statistically independent} \right\}.  
\end{split}
\end{equation}
For a given target dimension $R$, the set $\mathcal{U}$ and $\mathcal{V}$ defined in (\ref{eq:fair_CCA}) may be empty, making the constrained CCA problem undefined due to overly stringent restrictions. These restrictions may fail under general functions $f$ and $g$, as counterexamples exist where the transformed $i$th observation and $z_i$ remain statistically linked. To address this, following \cite{kleindessner2023efficient}, we relax (\ref{eq:fair_CCA}) by expanding the set $\mathcal{U}$ and $\mathcal{V}$ in two ways: first, we restrict our goal to linear functions of the form 
\begin{equation}
f(\m{a})= \m{w}_1^\top\m{a}+b_1 \quad \text{and}\quad g(\m{a})= \m{w}_2^\top\m{a}+b_2
\end{equation}
where $\m{w}_1,\m{w}_2\in\mathbb{R}^R$ and $b_1, b_2 \in \mb{R}$; second, rather than requiring $f(\m{U}^\top\m{X}_i^\top)$ and $z_i$ to be independent, we only require the two variables to be uncorrelated, that is their covariance to be zero. The same requirement applies to $g(\m{V}^\top\m{Y}_i^\top)$ and $z_i$. This leaves us with the following problem:
%
%
\begin{equation}\label{eq:fair_CCA_relaxed}
\begin{split}
\max_{\m{U},\m{V}}  \quad & \trace\left(\m{U}^\top\m{X}^\top\m{Y}\m{V}\right) \qquad
\text{subject to}  \quad \m{U}\in\mathcal{U}', \m{V}\in\mathcal{V}', \\
\text{where} \ 
&\mathcal{U}':=\left\{\m{U}\in\mathbb{R}^{D_\m{x}\times R}\mid\m{U}^\top\m{X}^\top\m{X}\m{U}=\m{I}_R;\right. \\  
&\left. \quad \forall \m{w}_1\in\mathbb{R}^R,b_1\in\mathbb{R}, \text{Cov}(\m{w}_1^\top\m{U}^\top\m{X}_i^\top+b_1,z_i)=0\right\}, \\
&\mathcal{V}':=\left\{\m{V}\in\mathbb{R}^{D_\m{y}\times R}\mid\m{V}^\top\m{Y}^\top\m{Y}\m{V}=\m{I}_R;\right. \\ 
& \left.\quad \forall \m{w}_2\in\mathbb{R}^R,b_2\in\mathbb{R},\text{Cov}(\m{w}_2^\top\m{V}^\top\m{Y}_i^\top+b_2,z_i)=0\right\}.
\end{split}
\end{equation}
Here, the ``$\text{Cov}(\m{x},\m{y})$'' refers to the covariance between $\m{x}\ \text{and}\ \m{y}$.

It can be seen that Problem (\ref{eq:fair_CCA_relaxed}) is well-defined. The optimization can be solved analytically similarly to standard CCA: with $\bar{z}=\frac{1}{N} \sum_{i=1}^N z_i$ and $\hat{\m{z}}=(z_1-\bar{z},\ldots, z_N-\bar{z})^\top \in\mathbb{R}^{N}$ such that 
\begin{equation*}
\centering
\begin{aligned}
&\forall \m{w}_1\in\mathbb{R}^R,b_1\in\mathbb{R}: \text{$\m{w}_1^\top\m{U}^\top\m{X}_i^\top+b_1$ and $z_i$ are uncorrelated}  \\
&\quad\Leftrightarrow\forall \m{w}_1\in\mathbb{R}^R,b_1\in\mathbb{R}: 
\sum_{i=1}^N (z_i-\bar{z})\cdot(\m{w}_1^\top\m{U}^\top\m{X}_i^\top+b_1)=0 \\
&\quad\Leftrightarrow\forall \m{w}_1: \m{w}_1^\top \m{U}^\top\m{X}^\top\hat{\m{z}}=0~\Leftrightarrow~ 
\m{U}^\top\m{X}^\top\hat{\m{z}}=\m{0},
\end{aligned}
\end{equation*}
where we ignore $b_1$ since it is arbitrary and independent of $\hat{\m{z}}$.
Similarly we have $\m{V}^\top\m{Y}^\top\hat{\m{z}}=\m{0}.$
We assume that $\m{X}^\top\hat{\m{z}}\neq \m{0}$ and $\m{Y}^\top\hat{\m{z}}\neq \m{0}$. Let $\m{R}_\m{x}\in\mathbb{R}^{D_\m{x}\times (D_\m{x}-1)}$ and $\m{R}_\m{y}\in\mathbb{R}^{D_\m{y}\times (D_\m{y}-1)}$ comprise as columns an orthonormal basis of the nullspace of $\m{X}^\top\m{z}$ and $\m{Y}^\top\m{z}$, respectively. 
Every $\m{U}\in\mathcal{U}'$ and $\m{V}\in\mathcal{V}'$ can then be written as 
\begin{equation}
\m{U}=\m{R}_\m{x}\g{\Lambda}_\m{x}\ \text{and} \ \m{V}=\m{R}_\m{y}\g{\Lambda}_\m{y}
\end{equation} 
where $\g{\Lambda}_\m{x}\in\mathbb{R}^{(D_\m{x}-1)\times R}$ with $\g{\Lambda}_\m{x}^\top\g{\Lambda}_\m{x}=\m{I}_R$ and $\g{\Lambda}_\m{y}\in\mathbb{R}^{(D_\m{y}-1)\times R}$ with $\g{\Lambda}_\m{y}^\top\g{\Lambda}_\m{y}=\m{I}_R$. Then the objective in (\ref{eq:fair_CCA_relaxed}) becomes 
\begin{equation}
\label{eq:new objective}
\trace\left(\g{\Lambda}_\m{x}^\top\m{R}_\m{x}^\top\m{X}^\top\m{Y}\m{R}_\m{y}\g{\Lambda}_\m{y}\right),
\end{equation}
where we now maximize w.r.t.~$\g{\Lambda}_\m{x}$ and $\g{\Lambda}_\m{y}$. The latter problem has exactly the form of CCA with $\m{X}$ replaced by $\m{XR}_\m{x}$ and $\m{Y}$ replaced by $\m{YR}_\m{y}$.
Once we have $\g{\Lambda}_\m{x}$ and $\g{\Lambda}_\m{y}$, we obtain a solution $\m{U}$ and $\m{V}$ of \eqref{eq:fair_CCA_relaxed} by computing $\m{U}=\m{R}_\m{x}\g{\Lambda}_\m{x}$ and $\m{V}=\m{R}_\m{y}\g{\Lambda}_\m{y}$. The final optimization problem of FR-CCA is to find $\m{U} \in  \mb{R}^{D_\m{x} \times R}$ and $\m{V} \in  \mb{R}^{D_\m{y} \times R }$ such that
\begin{equation}\label{eqn:main:FR-CCA}\tag{FR-CCA}
\begin{aligned}
\max & \quad \trace \left(\m{U}^\top\m{X}^\top\m{Y}\m{V}\right)  \\
\textnormal{subject to} & \quad 
\m{U}^\top\m{X}^\top \m{X}\m{U}=\m{V}^\top\m{Y}^\top\m{Y}\m{V} = \m{I}_R,\\
& \quad \m{U}^\top\m{X}^\top\hat{\m{z}}=\m{V}^\top\m{Y}^\top\hat{\m{z}}=\m{0}.
\end{aligned}
\end{equation}

After we have the new fair representations for the two modalities, i.e. $\m{X}_\text{fair}=\m{X}\m{U}$ and $\m{Y}_\text{fair}=\m{Y}\m{V}$, we can easily train classifiers on both representations with the same label $\m{y}$. The detailed procedure is summarized in Algorithm~\ref{alg:FR-CCA}.

\begin{algorithm}[t]
\small
\caption{FR-CCA and Subsequent Classification}
\label{alg:FR-CCA}
\begin{algorithmic}[1]
\Require Modalities $\m{X} \in \mathbb{R}^{N\times D_{\m{x}}}$, $\m{Y} \in \mathbb{R}^{N\times D_{\m{y}}}$; Sensitive attribute $\m{z} \in \mathbb{R}^N$; Label $\m{y} \in \mathbb{R}^N$
\Ensure Projected modalities $\m{X}_{\text{fair}}$, $\m{Y}_{\text{fair}}$; Classification results

\Statex \hspace{-\algorithmicindent} \textbf{Part 1: Learn Fair Representation}
\State Standardize $\m{X}$, $\m{Y}$, and $\m{z}$

\State Perform SVD on $\m{X}^\top \m{z}$: $[U_{\m{x}}, \Sigma_{\m{x}}, V_{\m{x}}^\top] = \text{SVD}(\m{X}^\top \m{z})$ 

\State Perform SVD on $\m{Y}^\top \m{z}$: $[U_{\m{y}}, \Sigma_{\m{y}}, V_{\m{y}}^\top] = \text{SVD}(\m{Y}^\top \m{z})$

\State Set $\m{R}_{\m{x}} = U_{\m{x}}^\top[:, 1:]$ and $\m{R}_{\m{y}} = U_{\m{y}}^\top[:, 1:]$

\State Project data to null spaces: $\m{X}_{\text{new}} = \m{X} \m{R}_{\m{x}}$, $\m{Y}_{\text{new}} = \m{Y} \m{R}_{\m{y}}$

\State Fit CCA on projected data: $\m{\Lambda}_{\m{x}}, \m{\Lambda}_{\m{y}} = \text{CCA}(\m{X}_{\text{new}}, \m{Y}_{\text{new}})$ 

\State Calculate $\m{U} = \m{R}_{\m{x}} \m{\Lambda}_{\m{x}}, \m{V} = \m{R}_{\m{y}} \m{\Lambda}_{\m{y}}$

\State Transform data: $\m{X}_{\text{fair}} = \m{X} \m{U}$, $\m{Y}_{\text{fair}} = \m{Y} \m{V}$

\Statex \hspace{-\algorithmicindent} \textbf{Part 2: Subsequent Classification}

\State Train a classifier (e.g., SVM) on $\m{X}_{\text{fair}}$ with labels $\m{y}$
\State Train a classifier (e.g., SVM) on $\m{Y}_{\text{fair}}$ with labels $\m{y}$
\State Compute fairness metrics: EOG, GSG, DPG

\end{algorithmic}
\end{algorithm}

\section{Experiments}
\begin{table*}[t]
\small
\renewcommand{\arraystretch}{0.82}
\caption{\label{tab:num}\small Numerical results (mean$\pm$std) in terms of three fairness metrics of SVM, CCA, SF-CCA, MF-CCA and our FR-CCA (the smaller the better). The best one for each modality in each line among the five methods is in bold.}
\begin{tabular}{c|l|l|l|l|l|l}
\toprule
\textbf{Modality}                & \makecell[c]{\textbf{Metric}} & \multicolumn{1}{c|}{\textbf{SVM}} & \multicolumn{1}{c|}{\textbf{CCA}} & \multicolumn{1}{c|}{\textbf{SF-CCA}} & \multicolumn{1}{c|}{\textbf{MF-CCA}} & \multicolumn{1}{c}{\textbf{FR-CCA}} \\ \midrule
\multirow{6}{*}{\textbf{Synthetic X}} & GSG           & $0.303\pm0.009$                   & $0.278\pm0.023$                   & $0.301\pm0.015$                      & $0.307\pm0.026$                      & $\m{0.242\pm0.046}$   \\
                                      & DPG           & $0.102\pm0.047$                   & $0.077\pm0.056$                   & $0.089\pm0.022$                      & $0.099\pm0.050$                      & $\m{0.026\pm0.031}$   \\
                                      & EOG           & $0.103\pm0.127$                   & $0.064\pm0.049$                   & $0.077\pm0.070$                      & $0.114\pm0.140$                      & $\m{0.012\pm0.014}$   \\
                                      & Precision     & $0.600\pm0.016$                   & $0.591\pm0.019$                   & $0.606\pm0.045$                      & $0.612\pm0.028$                      & $\m{0.635\pm0.022}$   \\
                                      & Recall        & $0.495\pm0.020$                   & $0.457\pm0.083$                   & $0.508\pm0.084$                      & $0.534\pm0.083$                      & $\m{0.599\pm0.029}$   \\
                                      & ROC-AUC Score & $0.493\pm0.017$                   & $0.486\pm0.017$                   & $0.479\pm0.048$                      & $\m{0.536\pm0.049}$             & $0.532\pm0.032$            \\ \midrule
\multirow{6}{*}{\textbf{Synthetic Y}} & GSG           & $0.274\pm0.038$                   & $0.329\pm0.042$                   & $0.286\pm0.009$                      & $0.296\pm0.004$                      & $\m{0.216\pm0.084}$   \\
                                      & DPG           & $0.075\pm0.069$                   & $0.081\pm0.040$                   & $0.100\pm0.069$                      & $0.104\pm0.057$                      & $\m{0.037\pm0.035}$   \\
                                      & EOG           & $0.107\pm0.131$                   & $0.084\pm0.037$                   & $0.077\pm0.076$                      & $0.095\pm0.117$                      & $\m{0.022\pm0.010}$   \\
                                      & Precision     & $0.624\pm0.026$                   & $0.612\pm0.033$                   & $0.632\pm0.038$                      & $\m{0.645\pm0.029}$             & $0.596\pm0.054$            \\
                                      & Recall        & $0.523\pm0.069$                   & $\m{0.612\pm0.033}$          & $0.477\pm0.073$                      & $0.512\pm0.053$                      & $0.569\pm0.017$            \\
                                      & ROC-AUC Score & $0.519\pm0.030$                   & $\m{0.811\pm0.016}$          & $0.508\pm0.084$                      & $0.537\pm0.030$                      & $0.525\pm0.037$            \\ \midrule
\multirow{6}{*}{\textbf{ADNI MRI}}    & GSG           & $0.329\pm0.042$                   & $0.302\pm0.230$                   & $0.286\pm0.033$                      & $0.296\pm0.004$                      & $\m{0.258\pm0.038}$   \\
                                      & DPG           & $0.081\pm0.040$                   & $0.026\pm0.032$                   & $0.036\pm0.024$                      & $0.104\pm0.057$                      & $\m{0.010\pm0.011}$   \\
                                      & EOG           & $0.084\pm0.037$                   & $0.020\pm0.018$                   & $0.033\pm0.031$                      & $0.095\pm0.117$                      & $\m{0.014\pm0.011}$   \\
                                      & Precision     & $0.612\pm0.033$                   & $0.552\pm0.627$                   & $0.619\pm0.026$                      & $\m{0.645\pm0.029}$             & $0.611\pm0.014$            \\
                                      & Recall        & $0.612\pm0.033$                   & $0.552\pm0.063$                   & $\m{0.619\pm0.026}$             & $0.512\pm0.053$                      & $0.611\pm0.014$            \\
                                      & ROC-AUC Score & $\m{0.811\pm0.016}$          & $0.782\pm0.017$                   & $0.800\pm0.020$                      & $0.537\pm0.030$                      & $0.773\pm0.027$            \\ \midrule
\multirow{6}{*}{\textbf{ADNI AV1451}} & GSG           & $0.329\pm0.042$                   & $0.239\pm0.208$                   & $0.286\pm0.033$                      & $0.242\pm0.042$                      & $\m{0.215\pm0.041}$   \\
                                      & DPG           & $0.081\pm0.040$                   & $0.028\pm0.024$                   & $0.036\pm0.024$                      & $0.015\pm0.013$                      & $\m{0.008\pm0.006}$   \\
                                      & EOG           & $0.084\pm0.037$                   & $0.015\pm0.017$                   & $0.033\pm0.031$                      & $0.026\pm0.022$                      & $\m{0.010\pm0.005}$   \\
                                      & Precision     & $0.612\pm0.033$                   & $\m{0.635\pm0.026}$          & $0.619\pm0.026$                      & $0.621\pm0.015$                      & $0.612\pm0.009$            \\
                                      & Recall        & $0.612\pm0.033$                   & $\m{0.635\pm0.026}$          & $0.619\pm0.026$                      & $0.621\pm0.015$                      & $0.612\pm0.009$            \\
                                      & ROC-AUC Score & $\m{0.811\pm0.016}$          & $0.791\pm0.020$                   & $0.800\pm0.020$                      & $0.769\pm0.024$                      & $0.798\pm0.020$            \\ \midrule
                                \textbf{Summary}   & Win Times & \makecell[c]{2} & \makecell[c]{4} & \makecell[c]{1} & \makecell[c]{3} & \makecell[c]{14} \\ \bottomrule
\end{tabular}
\end{table*}

\subsection{Datasets}

\begin{table}[t]
\centering
\small
\setlength{\tabcolsep}{5pt}
\caption{\small Demographic and diagnostic breakdown of the ADNI (MRI and AV1451) and synthetic datasets (X and Y).
}
\begin{tabular}{@{}clcccc|ccc@{}}
\toprule
\multicolumn{2}{c|}{\textbf{Dataset}}                         & \multicolumn{4}{c|}{\textbf{ADNI: MRI vs. AV1451}}                       & \multicolumn{3}{c}{\textbf{Synthetic: X vs. Y}}     \\ \midrule
\multicolumn{2}{l|}{\multirow{2}{*}{\textbf{\makecell{Diagnosis/\\Label}}}} & CN               & MCI             & AD               & Total            & Positive          & Negative           & Total           \\
\multicolumn{2}{l|}{}                                          & $224$            & $121$           & $30$             & $375$            & $304$             & $196$              & $500$           \\ \midrule
\multicolumn{2}{c|}{\multirow{2}{*}{\textbf{\makecell{Sensitive \\Attribute}}}}          & \multicolumn{2}{c}{Male}           & \multicolumn{2}{c|}{Female}         & Group 1           & \multicolumn{2}{c}{Group 2} \\
\multicolumn{2}{c|}{}                                          & \multicolumn{2}{c}{\textbf{$193$}} & \multicolumn{2}{c|}{\textbf{$182$}} & $398$             & \multicolumn{2}{c}{$102$}            \\ \bottomrule
\end{tabular}
\label{table:dataset_distribution}
\end{table}

\subsubsection{Synthetic Data} Following \cite{bach2005probabilistic,min2020tensor}, we employ multivariate Gaussian distribution to generate the data matrices $\m{X}\in\mathbb{R}^{N\times D_\m{x}}$ and $\m{Y}\in\mathbb{R}^{N\times D_\m{y}}$, integrating strong relationships with the sensitive attribute $\m{z}\in \mathbb{R}^{N}$. To achieve this, we take the exponential of a linear combination of odd-indexed and even-indexed features from $\m{X}$ and $\m{Y}$, subsequently blending these exponentiated components to form $\m{z}$. Then, the mean of $\m{z}$ across all samples is considered as the threshold to distinguish the sensitive subgroups. This process enhances the correlation between $\m{z}$ and the data features. For label generation, we combine the first half of the features from $\m{X}$, the latter half from $\m{Y}$, and the exponential of $\m{z}$. The mean of the aggregate serves as the binarization threshold, creating binary labels that simulate a potential sensitive bias within the dataset. The design mimics real-world scenarios where sensitive attributes may unfairly influence predictions, such as sex bias in Alzheimer's diagnosis. This synthetic data allows us to highlight the efficiency of FR-CCA in mitigating unfairness in classification tasks. 

\begin{figure*}[t]
\centering
\small 
\begin{minipage}{\linewidth}
\centering
\includegraphics[width=0.5\textwidth]{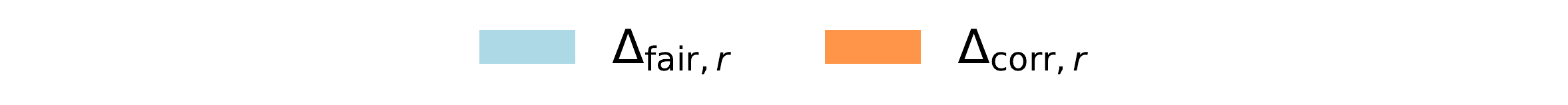}
\end{minipage}
\begin{minipage}{0.28\linewidth}
\centerline{\includegraphics[width=\textwidth]{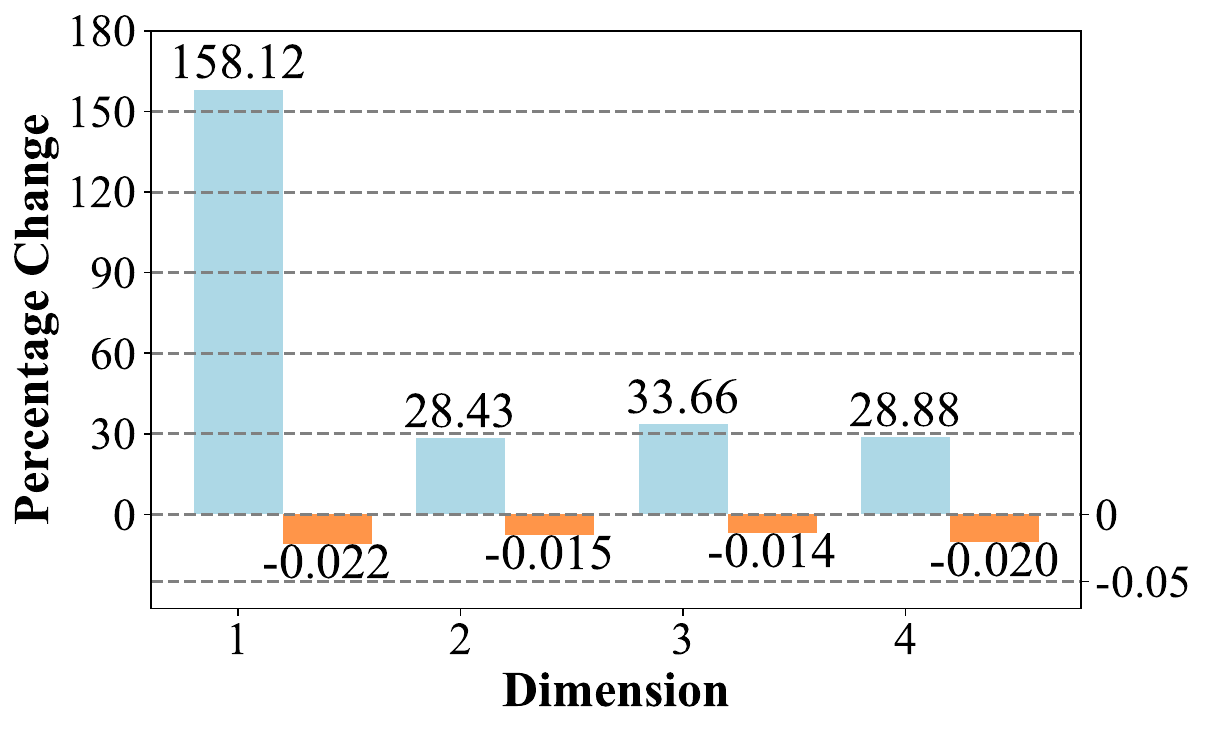}}
\centerline{ (a) FR-CCA v.s CCA}
\end{minipage}
\hspace{0.3cm}
\begin{minipage}{0.28\linewidth}
\centerline{\includegraphics[width=\textwidth]{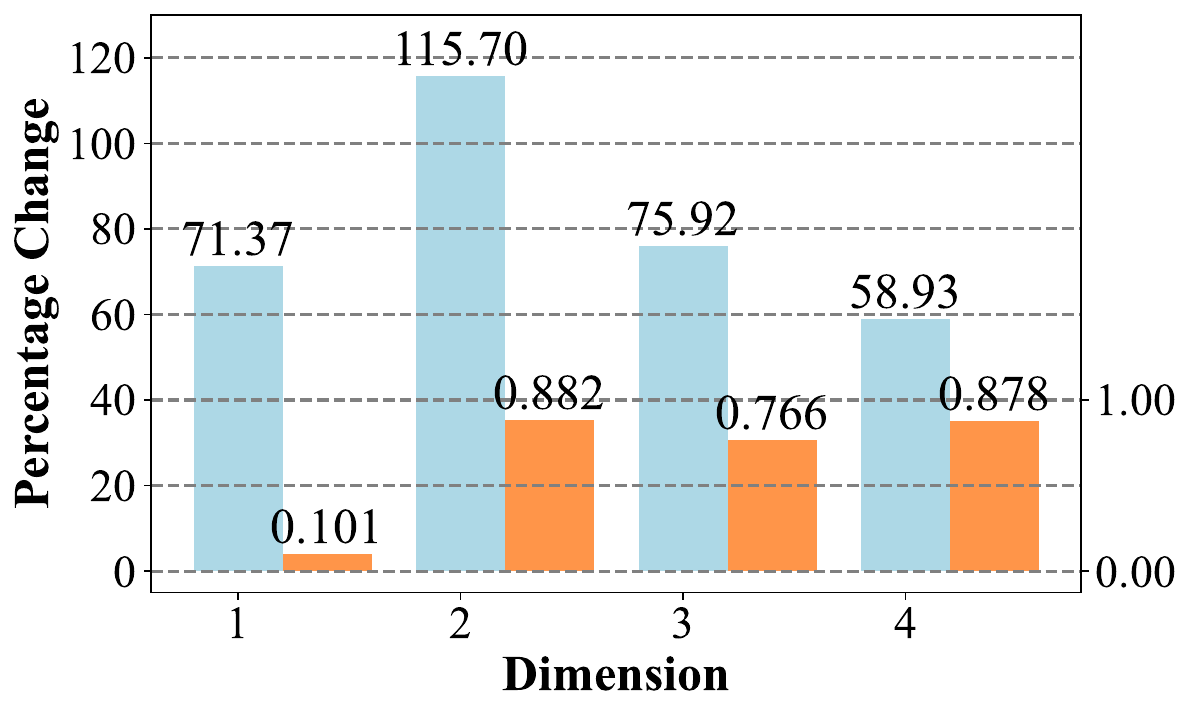}}
\centerline{ (b) FR-CCA v.s SF-CCA}
\end{minipage}
\hspace{0.3cm}
\begin{minipage}{0.28\linewidth}
\centerline{\includegraphics[width=\textwidth]{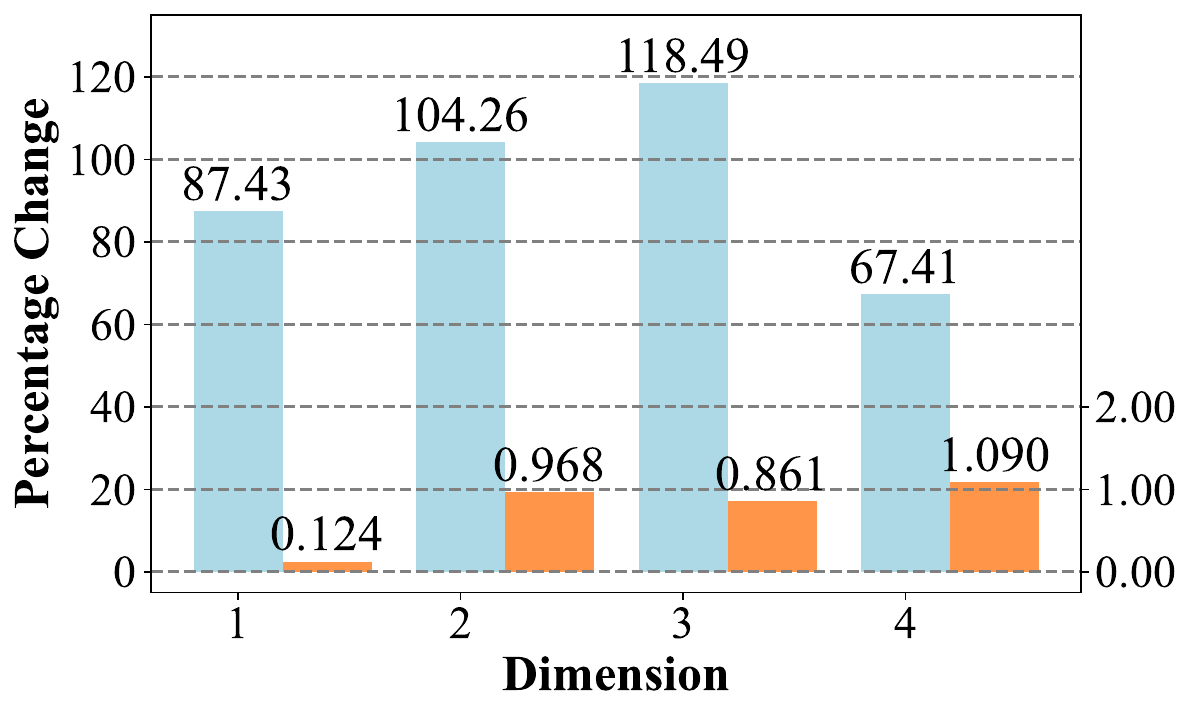}}
\centerline{ (c) FR-CCA v.s MF-CCA}
\end{minipage}
\begin{minipage}{0.28\linewidth}
\centerline{\includegraphics[width=\textwidth]{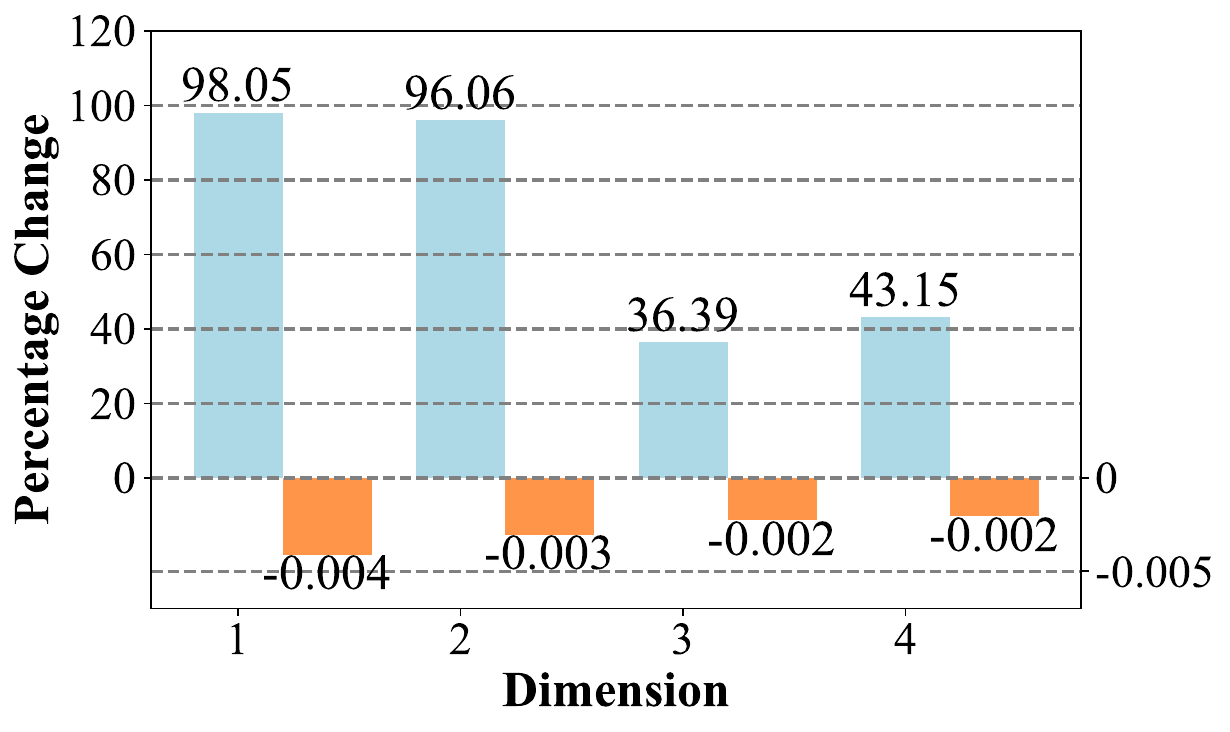}}
\centerline{(d) FR-CCA v.s CCA}
\end{minipage}
\hspace{0.3cm}
\begin{minipage}{0.28\linewidth}
\centerline{\includegraphics[width=\textwidth]{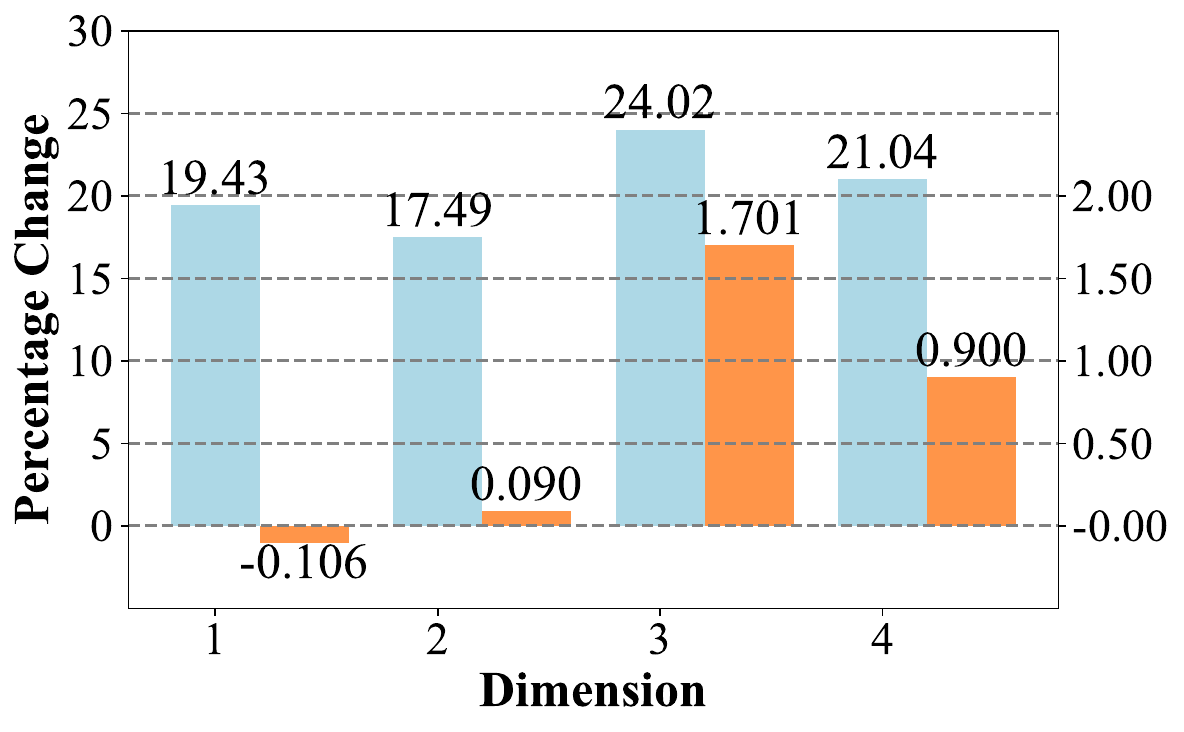}}
\centerline{(e) FR-CCA v.s SF-CCA}
\end{minipage}
\hspace{0.3cm}
\begin{minipage}{0.28\linewidth}
\centerline{\includegraphics[width=\textwidth]{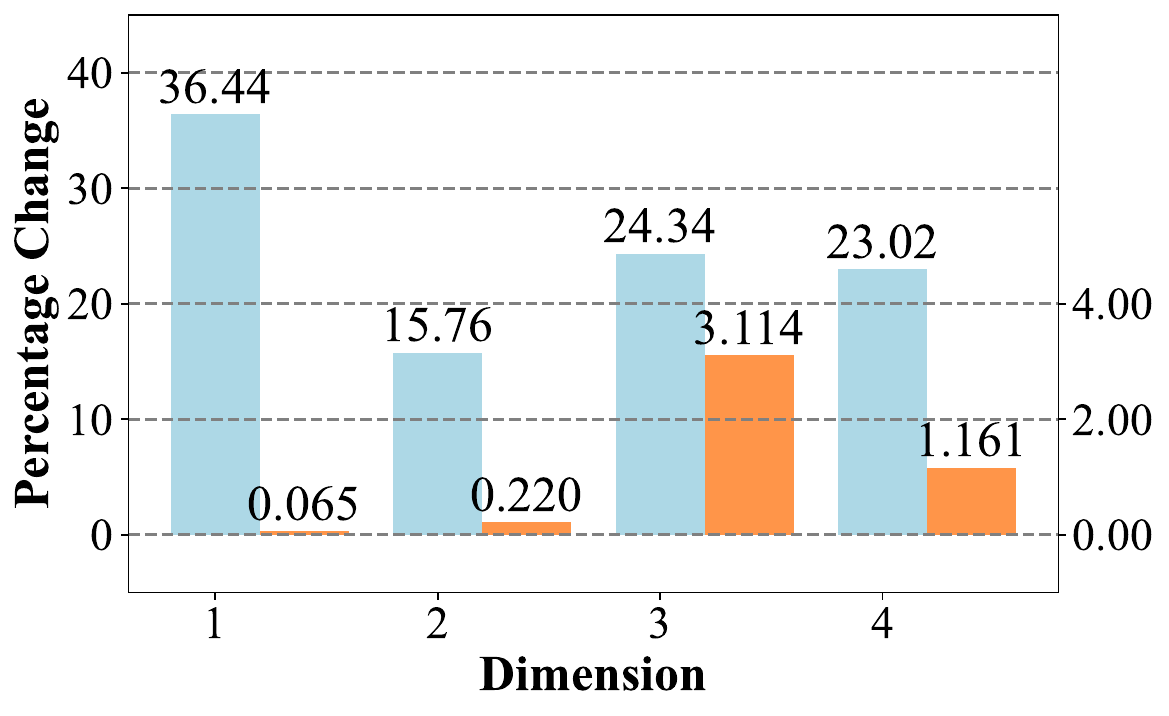}}
\centerline{(f) FR-CCA v.s MF-CCA}
\end{minipage}
\caption{\label{fig:CCA comparison} \small Percentage change of the fairness and correlation of our method compared to baseline methods. The \textbf{top row} (a–c) reports results on \textbf{synthetic data}, and the \textbf{bottom row} (d–f) on \textbf{ADNI data}. The x-axis is the projection dimension $r$. Blue columns ($\Delta_{\mathrm{fair},r}$) are plotted against the \emph{left} y-axis on a $\mathbf{\log_{10}}$ scale, 
  while orange columns ($\Delta_{\mathrm{corr},r}$) use the \emph{right} y-axis on a linear scale. 
  Numbers above the columns report the percentage changes on their respective scales. (We expect high percentage change for fairness metric indicating the effectiveness of our method in terms of fairness and low percentage change on correlation demonstrating small or even no loss in terms of correlation metric.)}
\end{figure*}

The synthetic data are generated by employing a multivariate Gaussian distribution. This formulation is given as follows:
\begin{equation}
    \begin{pmatrix}
        \m{X} \\
        \m{Y}
    \end{pmatrix} 
    \sim \mathcal{N}
    \begin{pmatrix}
        \begin{bmatrix}
            \mu_\m{x}\\
            \mu_\m{y}
        \end{bmatrix}, & 
        \begin{bmatrix}
            \m{\Sigma_{x}}& \m{\Sigma_{xy}} \\
            \m{\Sigma_{yx}}& \m{\Sigma_y}
        \end{bmatrix}
    \end{pmatrix},
\end{equation} where $\mu_\m{x} \in \mathbb{R}^{D_\m{x}}$ and $\mu_\m{y} \in \mathbb{R}^{D_\m{y}}$ denote the mean vectors for the data matrices $\m{X}$ and $\m{Y}$, correspondingly; covariance matrices $\m{\Sigma_{x}}$, $\m{\Sigma_{y}}$ and the cross-covariance matrix $\m{\Sigma_{xy}}$ are constructed as follows.
Given orthogonal projection matrices $\m{U} \in \mathbb{R}^{D_\m{x} \times R},$ $\m{V} \in \mathbb{R}^{D_\m{y} \times R}$, from the Haar distribution, and canonical correlations $\g{\rho} = (\rho_1, \rho_2, \ldots, \rho_R)$ defined as $\g{\rho} = \text{diag}(\m{U^\top X^\top YV})$. Let $\m{U} = \m{Q}_\m{x}\m{R}_\m{x}$ and $\m{V} = \m{Q}_\m{y}\m{R}_\m{y}$ be the QR decomposition of $\m{U}$ and $\m{V}$, then we have
\begin{align*}
    \m{\Sigma_{xy}} &= \m{\Sigma_{x}U\text{diag}(\g{\rho})V^\top \Sigma_y}\\
    \m{\Sigma_x} &= \m{Q_x (R_x^{\top})^{+}R_x^{+} + \epsilon_x (I_{D_\m{x}}-Q_x Q_x^\top)} \\
    \m{\Sigma_y} &= \m{Q_y (R_y^{T})^{+}R_y^{+} + \epsilon_y (I_{D_\m{y}}-Q_y Q_y^\top)}
\end{align*}
Here, $\epsilon_x$ and $\epsilon_y$ denote the Gaussian noise levels; $\m{A}^{+}$ represents the Moore-Penrose inverse of matrix $\m{A}$.

To simulate the sex subgroups within the synthetic dataset, the sensitive attribute $\m{z}$ is constructed as follows: Let $\m{X}_\text{odd}$ and $\m{Y}_\text{even}$ represent the matrices containing the odd-indexed features of $\m{X}$ and the even-indexed features of $\m{Y}$ respectively. We define the sensitive attribute value of sample $i$ by:
\begin{equation}
    z_i = \alpha\exp\left(\sum_{j}a_j \m{X}_{\text{odd}_{ij}}\right) + \beta\exp\left(\sum_{k}b_k\m{Y}_{\text{even}_{ik}}\right),
\end{equation}
where $\alpha$ and $\beta$ are weights determining the contribution of each modality to the sensitive attribute, and $a_j, b_k$ are the coefficients of the linear combinations. The attribute $\m{z}$ is then binarized based on:
\begin{equation}
    z_i = \begin{cases}
        1 & \text{ if }z_i \leq \tau \\
        2 & \text{ if }z_i > \tau
    \end{cases}, ~\text{where}~\tau = \frac{1}{N}\sum_{i=1}^Nz_i.
\end{equation}
This binarization splits the dataset into two subgroups, intended to simulate sex divisions, and is utilized to assess the model's fairness in handling sex-related biases.

To generate the labels $\m{y} \in \mathbb{R}^N$, we first define the combined feature vector $\m{c} \in \mathbb{R}^N$ as: 
\begin{equation}
    \m{c} =\sum_{i = 1}^{\lfloor\frac{D_\m{x}}{2}\rfloor}\m{X}_{:,i} + \sum_{j=\lfloor\frac{D_\m{y}}{2}\rfloor}^{D_\m{y}} \m{Y}_{:,j} + \exp{(\m{z})}
\end{equation}
where $\m{X}_{:,i}, \m{Y}_{:,i}$ represents the $i$th column of $\m{X},\m{Y}$ respectively and $\lfloor\cdot\rfloor$ denotes the floor function which returns the greatest integer less than or equal to its argument. The binary label for each instance is then given by:
\begin{equation}
    y_i = \begin{cases}
        1 & \text{ if } c_i \leq t \\
        2 & \text{ if } c_i > t\\
    \end{cases}, ~\text{where}~t =\frac{1}{N}\sum_{i=1}^Nc_i.
\end{equation}

As shown in Table~\ref{table:dataset_distribution}, our synthetic datasets refer to two modalities $\m{X}$ and $\m{Y}$, where $\m{X}$ contains $55$ features and $\m{Y}$ comprises $60$ features. These datasets include $500$ samples divided into $304$ positive and $196$ negative cases. The distribution of participants into two sensitive groups, with $398$ in Group 1 and $102$ in Group 2, allows for comprehensive modeling and validation of diagnostic algorithms. The synthetic datasets provide a controlled environment to evaluate the robustness and generalizability of predictive models developed using the ADNI datasets.

\subsubsection{Real Data} Our experiment incorporates two medical imaging modalities: (1) Magnetic Resonance Imaging (MRI) scans and (2) Tau (AV1451) positron emission tomography (PET) scans from the Alzheimer’s Disease Neuroimaging Initiative (ADNI) database \cite{weiner2013aad,weiner2017recent}. The ADNI was launched in 2003 as a public-private partnership led by Principal Investigator Michael W. Weiner, MD. ADNI aims to determine if serial magnetic resonance imaging (MRI), positron emission tomography (PET), other biological markers, and clinical and neuropsychological assessments can be integrated to track the progression of MCI and early AD. (\url{http://adni.loni.usc.edu})

Table~\ref{table:dataset_distribution} shows that the dataset comprises 375 subjects (182 males, 193 females), with 66 cortical thickness features from MRI and 68 cortical standardized uptake value ratio (SUVR) features measuring tau accumulation level from AV1451 scans. Label distribution 
is categorized into cognitive normal (CN, N=224), mild cognitive impairment (MCI, N=121), and Alzheimer’s Disease subjects (AD, N=30). ADNI data are analyzed to explore fair representation through FR-CCA and its impact on subsequent medical imaging classification tasks, with a focus on sex-based subgroup fairness.

\begin{figure*}[t]
  \centering
  \begin{minipage}{1\textwidth}
  \centering
    \includegraphics[width=1\textwidth]{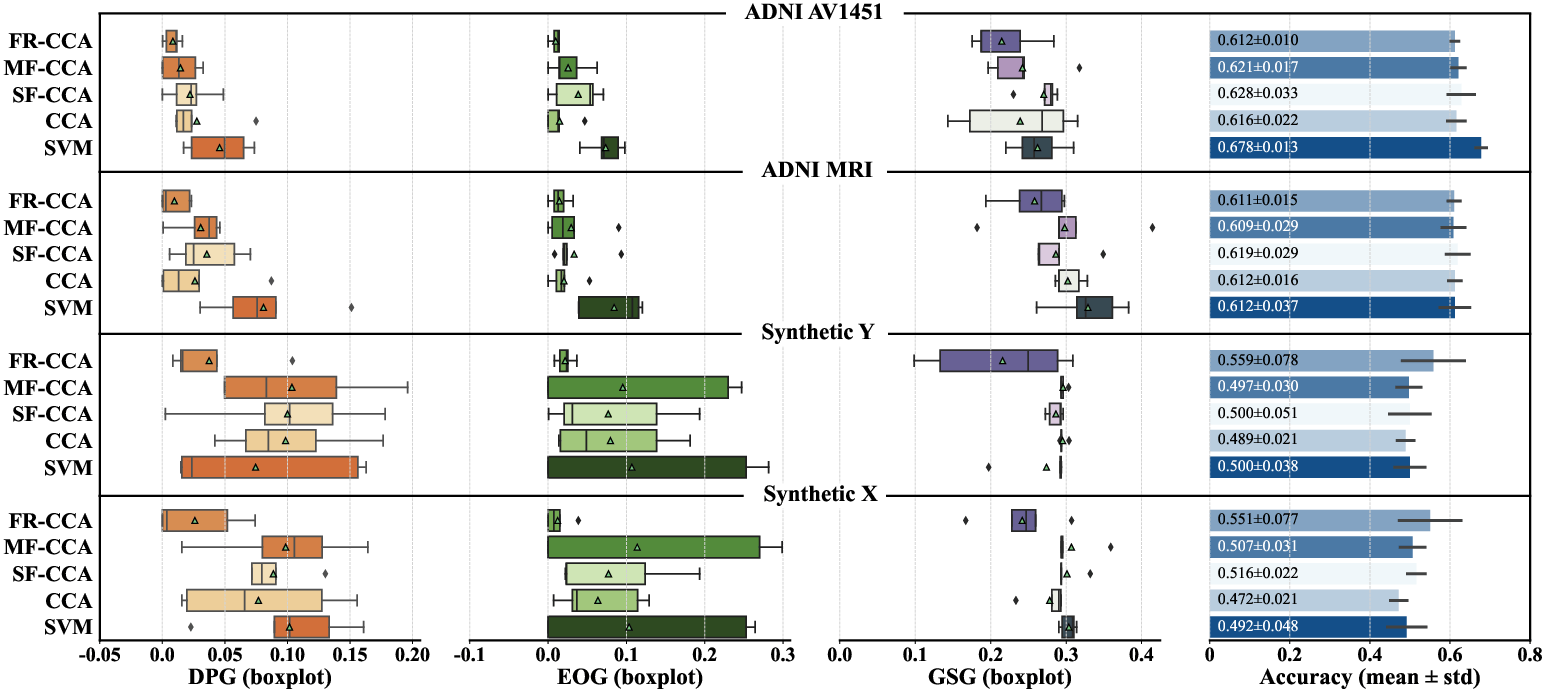}
  \end{minipage}
  \caption{\label{fig:classification}\small Comparison plot for fairness metrics (DPG, EOG, GSG) and Accuracy on four modalities (X and Y for synthetic data, MRI and AV1451 for ADNI data). We use boxplots to illustrate the results for three fairness metrics as they exhibit significant variation across different methods. We use boxplots to display the accuracy, as there is minimal variation across different methods. The green triangle represents the mean value over five runs in the boxplots, and the dark diamonds refer to outliers. Our FR-CCA outperforms all the baseline models regarding three fairness metrics (the smaller, the better) while also achieving competitive accuracy results (the larger, the better).}
\end{figure*}

\subsection{Setting}
Our experimental framework involves two stages: an unsupervised learning phase focused on discovering a meaningful representation, followed by a subsequent classification task.
In the unsupervised learning stage, FR-CCA's performance is evaluated on the percentage change of correlation and fairness compared to CCA, SF-CCA, and MF-CCA for $r$th dimension of representation spaces, where $ r= 1\ldots, R$. Let $\m{U} = [\m{u}_1, \ldots, \m{u}_R]\in \mathbb{R}^{D_\m{x} \times R}$ and $\m{V} = [\m{v}_1, \ldots, \m{v}_R] \in \mathbb{R}^{D_\m{y}\times R}$, and then the $r$th canonical correlation, is formulated as:
\begin{equation}
    \boldsymbol{\rho}_r = \frac{\m{u}_r^\top \m{X}^\top \m{Y} \m{v}_r}{\sqrt{\m{u}_r^\top \m{X}^\top \m{X}\m{u}_r\m{v}_r^\top \m{Y}^\top\m{Y}\m{v}_r}}.
\end{equation}
Then, the percentage change of correlation is given by:
\begin{equation}
    \Delta_{\text{corr}, r} = \frac{\boldsymbol{\rho}^{(\text{p})}_r - \boldsymbol{\rho}^{(\text{b})}_r}{\boldsymbol{\rho}^{(\text{b})}_r} \times 100\%,
\end{equation}
where ``(p)'' refers to the proposed method and ``(b)'' refers to a baseline method.
In this unsupervised scenario, we define 
\begin{equation}
\label{eq:orthogonal}
\m{\boldsymbol{\gamma}_x = U^\top X^\top \hat{z}, \quad \boldsymbol{\gamma}_y = V^\top Y^\top \hat{z}}
\end{equation} in $\m{X}, \m{Y}$ respectively (refer to \eqref{eqn:main:FR-CCA}), and then the fairness metric for $r$th dimension in CCA task (unsupervised learning phase) is defined by 
\begin{equation}
\boldsymbol{\gamma}_r=\frac{[\boldsymbol{\gamma}_\m{x}]_r+[\boldsymbol{\gamma}_\m{y}]_r}{2}.
\end{equation}
Then, the percentage change of fairness is given by:
\begin{equation}
\label{eq:percentage change}
    \Delta_{\text{fair}, r} = -\frac{\boldsymbol{\gamma}_r^{(\text{p})}-\boldsymbol{\gamma}_r^{(\text{b})}}{\boldsymbol{\gamma}_r^{(\text{b})}} \times 100 \%.
\end{equation}

In the subsequent classification task, we employ support vector machine (SVM)~\cite{steinwart2008support} for classification, which is robust and effective in handling datasets with complicated structures. We also train logistic regression (LR) and multilayer perceptron (MLP) models on these fair representations and put their results in the Supplementary Material. To assess the performance of the classifier, we focus on four metrics (refer to Section \ref{sec:pre}): Demographic Parity Gap (DPG), Equalized Odds Gap (EOG), Group Sufficiency Gap (GSG), and Accuracy. A discussion on why we choose these fairness metrics can be found in the Supplementary Material.

For hyperparameter tuning, a specific random seed is allocated for optimal parameter selection. The dataset is divided into a training set (70\%) and a testing set (30\%). Stratified 5-fold cross-validation is applied to the training dataset. Subsequently, a random search is conducted using three distinct scorers-DPG, EOG, and GSG-each is designed to measure the model's performance against fairness metrics. We utilize kernels such as Sigmoid and Radial Basis Function (RBF) due to the non-linear separability of the dataset and uniform distribution to search for the optimal parameters: the penalty parameter $C \in [0.1, 200]$, the kernel coefficient $\gamma \in [0.1, 200]$ along with ``scale'' and ``auto'' options, and the kernel coefficient $\text{coef}0 \in [0,50]$ (``Sigmoid'' only). 
In the actual training phase, we use five additional seeds to split the data into training (70\%) and testing (30\%) sets with the optimally tuned hyperparameters from the previous step to train the model. The final evaluation criteria are based on the mean and standard deviation of outcomes across these seeds, offering a comprehensive assessment of the model's fairness, accuracy, performance, and robustness. This methodology ensures a balanced evaluation that reflects the model's ability to generalize across different datasets while retaining fair and accurate predictions.


\subsection{Baselines}
To evaluate the effectiveness and fairness of our proposed FR-CCA, we compare its performance against two fairness-enhanced CCA methods and two foundational baselines. SF-CCA and MF-CCA~\cite{zhou2024fair} are included to benchmark fairness metrics. Foundational CCA~\cite{hotelling1992relations} serves as a baseline to demonstrate the contrast in maximizing correlation without the fairness enhancement. Additionally, an SVM classifier trained on the original data acts as a performance benchmark, highlighting the effects of preprocessing steps on raw prediction. This diverse set of baselines allows us to highlight FR-CCA's ability to improve fairness without sacrificing the correlation and predicting accuracy. We also compare our method with methods that focus solely on fair classification and the results can be found in the Supplementary Material. The discussion on the key difference between our FR-CCA and F-CCA can also be found in the Supplementary Material.

\subsection{Results}

\subsubsection{Results on synthetic data}
In the simulation experiment, we follow the methodology demonstrated in Section 4.1 to generate two synthetic datasets $\m{X}\in \mathbb{R}^{N \times D_\m{x}}$, $\m{Y}\in \mathbb{R}^{N \times D_\m{y}}$ with $N = 500, D_\m{x} = 55, D_\m{y} = 60$, along with sensitive attributes $\m{z}$ and labels by the given ground truth projection matrices $\m{U}\in \mathbb{R}^{D_\m{x} \times R}, \m{V}\in\mathbb{R}^{D_\m{y} \times R}$, and canonical correlation $\boldsymbol{\rho} = [0.8, 0.6, 0.3, 0.5]$.

The unsupervised learning stage compares FR-CCA with three baselines: CCA, SF-CCA, and MF-CCA. Figure~\ref{fig:CCA comparison}(a,b,c) shows the percentage change in fairness and correlation when using FR-CCA compared to the baselines across various dimensions. FR-CCA demonstrates significant improvements in fairness, with slight decreases or even no loss compared to all three baselines. Note that here the percentage change defined in (\ref{eq:percentage change}) will be very close to 100\% because $\boldsymbol{\gamma}_\m{x}$ and $\boldsymbol{\gamma}_\m{y}$ in (\ref{eq:orthogonal}) are all close to 0 ((smaller than $10^{-10}$) ) guaranteed by our method. To illustrate the order-of-magnitude differences across different dimensions, we use $\mathbf{\log_{10}}$ to scale $\boldsymbol{\gamma}_\m{x}$ and $\boldsymbol{\gamma}_\m{y}$ before calculating the percentage change.

We run the experiment on $R = 2$ dimension in the subsequent classification task. As shown in Figure~\ref{fig:classification} (bottom two lines), FR-CCA significantly outperforms other algorithms in reducing DPG, EOG, and GSG, suggesting an obvious mitigation of bias across distinct sensitive subgroups while keeping competitive accuracy, illustrating a balanced trade-off between fairness and accuracy in the prediction task.
FR-CCA performs well on both Synthetic X and Synthetic Y modalities by achieving high precision, recall, and ROC-AUC scores, demonstrating its capability to match other methods in traditional performance metrics while significantly enhancing fairness. This balance of performance and fairness makes FR-CCA a superior choice for equitable and accurate prediction. The numerical values (mean$\pm$std) of the three fairness metrics and accuracy are presented in Table~\ref{tab:num} (first two rows) and Figure~\ref{fig:classification} (first two rows, last column), respectively. The discussion on the choice of $R$ can be found in the Supplementary Material.

\subsubsection{Results on real data} The analysis of ADNI data strictly follows the steps in synthetic data. As depicted in Figure~\ref{fig:CCA comparison}(d,e,f), FR-CCA remains a significant improvement of fairness with almost no loss of correlation compared to other baselines across various dimensions in the unsupervised learning phase.

%

We apply the subsequent classification tasks on MRI and AV1451 modalities separately. As illustrated in Figure~\ref{fig:classification} (top two rows), our approach not only preserves the predicting accuracy but also significantly reduces the sex-biased prediction. Low GSG, DPG, and EOG values are crucial in Alzheimer's disease diagnosis as they indicate minimal bias and high fairness across different demographic groups. This ensures that the diagnostic tool provides equitable and accurate patient assessments, leading to more consistent and reliable diagnoses. Reducing these gaps helps prevent misdiagnosis and underdiagnosis in historically disadvantaged populations, ultimately supporting better, more inclusive healthcare outcomes. Additionally, the standard deviation of results by FR-CCA across distinct random seeds is notably small, indicating FR-CCA's superior robustness in classification tasks. 

The overall performance of FR-CCA on both the ADNI MRI and ADNI AV1451 datasets suggests that it is a highly effective method for diagnosing Alzheimer's disease. Its strength lies in its ability to maintain a balanced and high performance across precision and recall, which is essential for clinical applicability. By effectively minimizing both false positives and false negatives, FR-CCA ensures that patients receive accurate diagnoses, which is critical for timely and appropriate treatment interventions.

The consistently high ROC-AUC scores across different modalities indicate that FR-CCA is reliable and capable of generalizing well, making it a valuable tool in the clinical setting. This method's ability to provide robust performance metrics highlights its potential for integration into diagnostic workflows, ultimately contributing to improved patient outcomes in Alzheimer’s disease management.

Briefly, the balanced performance of FR-CCA across traditional metrics and fairness metrics suggests it is well-suited for integration into clinical workflows. The numerical values (mean$\pm$std) of the three fairness metrics and accuracy are presented in Table~\ref{tab:num} (last two rows) and Figure~\ref{fig:classification} (last two rows, last column), respectively. A trade-off discussion between accuracy and fairness can be found in the Supplementary Material.

\subsubsection{Interpretability Study}
\begin{figure*}[t]
\small
    \centering
    \begin{minipage}{0.45\linewidth}
    \centerline{\includegraphics[width=\textwidth]{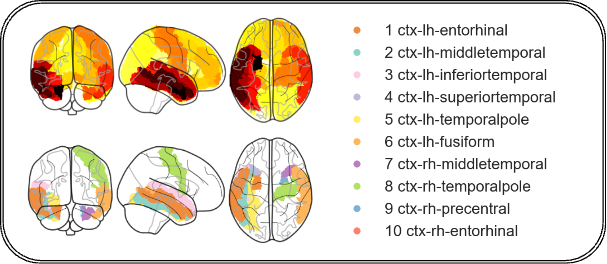}}
    \centerline{ (a) \textbf{ADNI MRI}}
    \end{minipage}
    \begin{minipage}{0.5\linewidth}
    \centerline{\includegraphics[width=\textwidth]{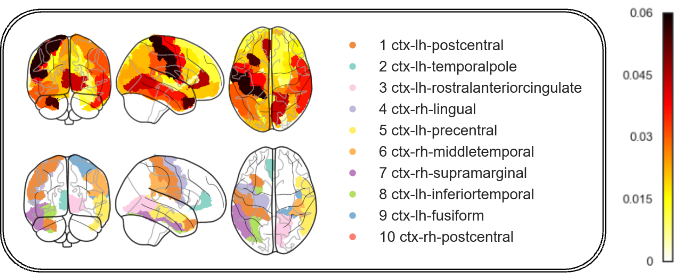}}
    \centerline{\hspace{-0.8cm} (b) \textbf{ADNI AV1451}}
    \end{minipage}
    \caption{\small Brain heat map of feature importance for ADNI MRI and ADNI AV1451 using the coefficient of our FR-CCA model. Each modality includes three slices to display the brain region extensively. The heat maps in the first row showcase all the brain regions of interest (ROI), where darker colors represent greater importance for risk prediction.
    The second row highlights the top ten significant brain regions in each modality, which are annotated in the legends in the right column.}
    \label{fig:feature_importance}
\end{figure*}
The subsequent prediction task for our FR-CCA model is based on an SVM with an RBF kernel. Unlike a linear SVM, we cannot directly extract the parameter coefficients from the classifier. Therefore, we employ SHAP (SHapley Additive exPlanatons) values~\cite{lundberg2017unified} to calculate the feature importance, which provides a unified measure of feature importance by interpreting the prediction of machine learning models in the low dimensional representation space. We then utilize the projection matrices $U \in \mathbb{R}^{68\times 2}$ and $V\in \mathbb{R}^{66\times 2}$ learned by FR-CCA to map the feature importance for the first two canonical components back to the original feature spaces, where $68$ brain regions for MRI and $66$ brain regions for AV1451. 
Figure~\ref{fig:feature_importance} shows the important brain regions that our model focuses on in the diagnosis for Alzheimer's disease. As can be seen, the first row shows the importance of AD risk in all the regions. The second row highlights the top ten important regions for each modality. 

For the MRI modality, focusing on neurodegeneration, the most critical regions include those involved in memory, language comprehension, and high-level visual processing. The entorhinal cortex is notably one of the earliest affected areas in AD~\cite{entorhinalAD}, correlating with significant memory impairment due to its role in memory and navigation. Temporal regions, including the middle and inferior temporal gyri, are vital for semantic memory and language comprehension, and their degeneration leads to comprehension difficulties and memory issues~\cite{temporal_region_Frisoni}. The degeneration of superior temporal gyrus contributes to language and auditory comprehension deficits~\cite{superior_temporal}. The temporal pole, associated with high-order cognitive functions and relaying information, also shows degeneration, leading to deficits in decision-making and working memory~\cite{temporalpole}. The fusiform gyrus, essential for high-level visual processing, exacerbates visual recognition deficits upon degeneration~\cite{fusiformgyrus}. Additionally, the precentral and postcentral gyri show progressive degeneration as AD advances, impacting motor functions and sensory integration~\cite{precentral}. 

For the AV1451 modality, which assesses tau pathology, the top regions of interest include areas critical for sensory processing, emotional regulation, decision-making, and visual processing. The postcentral gyrus is involved in processing sensory information, and tau accumulation here disrupts sensory perceptron and processing~\cite{Samrah2018}. The temporal pole is associated with emotional regulation and social cognition with tau accumulation. The rostral anterior cingulate is crucial for emotion regulation and decision-making, and tau pathology here leads to increased apathy and impaired decision-making~\cite{IACCARINO2018452}. The lingual gyrus, important for visual processing, contributes to visual processing deficits when affected by tau. The precentral gyrus, part of the primary motor cortex, leads to a motor dysfunction with tau pathology. The middle and inferior temporal gyri are vital for language and visual object recognition, and tau accumulation exacerbates language comprehension and visual recognition issues~\cite{temporal_region_Frisoni}. The supramarginal gyrus contributes to language processing deficits with tau accumulation. The heat map visually represents these critical regions, providing a comprehensive understanding of their contributions to AD progression and aiding in accurate disease diagnosis.
\subsubsection{Time Comparison}

\begin{table}[t]
\centering
\small
\setlength{\tabcolsep}{3pt}
\caption{\label{tab:time}\small Time (seconds, mean$\pm$ std) comparison over $10$ experiments across different methods ($R=5$ on the ADNI dataset and $R=7$ on the synthetic dataset). 
}
\begin{tabular}{l|c|c|c|c}
\toprule
\textbf{Dataset} & \textbf{CCA}        & \textbf{SF-CCA}     & \textbf{MF-CCA}      & \textbf{FR-CCA}      \\ \midrule
Synthetic         & $0.256 \pm 0.008$ & $7.322 \pm 1.202$ & $31.112 \pm 1.890$ & $0.346 \pm 0.011 $ \\ \midrule
ADNI             & $0.305 \pm 0.013$ & $8.694 \pm 1.574$ & $29.382 \pm 2.108$ & $0.504 \pm 0.016$  \\ \bottomrule
\end{tabular}
\end{table}

Furthermore, we compare the timing performance across CCA, SF-CCA, MF-CCA, and FR-CCA in the unsupervised learning phase shown in Table~\ref{tab:time}. Note that the experiments are run on $12th$ Gen Intel(R) Core(TM) i9-12900K 3.19GHz. FR-CCA achieves competitive timing performance compared to CCA, which is consistent with the time analysis in the Supplementary Material and significantly outperforms SF-CCA and MF-CCA, which demonstrates the efficiency and applicability of the proposed method.



\section{Discussion}
\label{sec:discussion}

FR-CCA provides an efficient closed-form solution to fair multimodal representation learning through nullspace projection. While our approach successfully removes linear dependencies between representations and sensitive attributes, achieving substantial improvements across fairness metrics (DPG, EOG, GSG), we acknowledge the theoretical trade-off in relaxing from statistical independence to uncorrelatedness. This design choice prioritizes computational tractability and clinical deployability over theoretical completeness.

The method naturally extends to multiple sensitive attributes by projecting onto $(D_x-k)$ and $(D_y-k)$ dimensional subspaces, though correlated attributes require careful handling to avoid excessive information loss. In biomedical applications like Alzheimer's disease diagnosis, where sensitive attributes (e.g., age, sex) are often entangled with legitimate predictive signals, our approach necessarily trades some accuracy for fairness—a fundamental challenge in fair representation learning.

Our evaluation using complementary group fairness metrics (DPG, EOG, GSG) aligns with healthcare contexts where group-level disparities are primary concerns. The application to ADNI data is particularly motivated by documented sex-based disparities in AD, with women having approximately twice the lifetime risk. FR-CCA maintains competitive accuracy while consistently achieving superior fairness across all metrics and modalities, demonstrating its effectiveness in balancing these often-competing objectives.

The computational efficiency (O$(D_x + D_y)^3$) and interpretability of our linear projections make FR-CCA particularly suitable for clinical deployment. Future work should explore stronger independence constraints and methods for handling correlated sensitive attributes without excessive information loss. For extended discussions on methodological trade-offs, comparisons with related methods, and implementation considerations, please refer to the Supplementary Material.

\section{Conclusion}

This paper presents a novel fair CCA method that addresses the previous limitations by focusing on fair representation learning. By ensuring the independence of projected features from sensitive attributes, our method achieves improved fairness performance, such as Group Sufficiency Gap (GSG), Demographic Parity Gap (DPG), and Equalized Odds Gap (EOG), without sacrificing accuracy in subsequent classification tasks. The empirical studies conducted on both synthetic data and ADNI data demonstrate the effectiveness of the proposed fair CCA method in maintaining high correlation analysis performance while simultaneously enhancing fairness. This novel method excels in traditional performance metrics such as precision, recall, and ROC-AUC scores, ensuring accurate and reliable diagnoses crucial for clinical applications. This balance of high performance and enhanced fairness makes FR-CCA a robust and reliable tool for Alzheimer's disease diagnosis, promoting equitable healthcare outcomes across diverse patient populations.


\section*{Acknowledgments}
We would like to thank Davoud Ataee Tarzanagh for helpful discussions. We would also like to thank all the anonymous reviewers for their constructive feedback to improve this paper. 

This work was supported in part by the NIH grants U01 AG066833, RF1 AG063481, U01 AG068057, R01 LM013463, R01 AG071470, U19 AG074879 and a PSOM
AI2D Seeding Project.
%


Data collection and sharing for this project was funded by the Alzheimer's Disease Neuroimaging Initiative (ADNI) (National Institutes of Health Grant U01 AG024904) and DOD ADNI (Department of Defense award number W81XWH-12-2-0012). ADNI is funded by the National Institute on Aging, the National Institute of Biomedical Imaging and Bioengineering, and through generous contributions from the following: AbbVie, Alzheimer's Association; Alzheimer's Drug Discovery Foundation; Araclon Biotech; BioClinica, Inc.; Biogen; Bristol-Myers Squibb Company; CereSpir, Inc.; Cogstate; Eisai Inc.; Elan Pharmaceuticals, Inc.; Eli Lilly and Company; EuroImmun; F. Hoffmann-La Roche Ltd and its affiliated company Genentech, Inc.; Fujirebio; GE Healthcare; IXICO Ltd.; Janssen Alzheimer Immunotherapy Research \& Development, LLC.; Johnson \& Johnson Pharmaceutical Research \& Development LLC.; Lumosity; Lundbeck; Merck \& Co., Inc.; Meso Scale Diagnostics, LLC.; NeuroRx Research; Neurotrack Technologies; Novartis Pharmaceuticals Corporation; Pfizer Inc.; Piramal Imaging; Servier; Takeda Pharmaceutical Company; and Transition Therapeutics. The Canadian Institutes of Health Research is providing funds to support ADNI clinical sites in Canada. Private sector contributions are facilitated by the Foundation for the National Institutes of Health (www.fnih.org). The grantee organization is the Northern California Institute for Research and Education, and the study is coordinated by the Alzheimer's Therapeutic Research Institute at the University of Southern California. ADNI data are disseminated by the Laboratory for Neuro Imaging at the University of Southern California. 


\balance
\bibliographystyle{ACM-Reference-Format}
\bibliography{ref}

\end{document}